\pdfoutput=1

\documentclass[11pt]{article}

\usepackage[]{acl2021}

\usepackage{times}
\usepackage{latexsym}

\usepackage[T1]{fontenc}

\usepackage[utf8]{inputenc}

\usepackage{microtype}

\usepackage{times}
\usepackage{latexsym}
\usepackage{graphicx}

\usepackage{microtype}

\aclfinalcopy 

\title{StoryDB: Broad Multi-language Narrative Dataset}

\author{
Alexey Tikhonov\\
Yandex\\
Berlin, Germany\\
\texttt{altsoph@gmail.com}\And
Igor Samenko\\
ICT RAS\\
Novosibirsk, Russia\And
Ivan P. Yamshchikov\\
LEYA Laboratory\\
Yandex and\\
Higher School of Economincs\\
St. Petersburg, Russia\\
\texttt{ivan@yamshchikov.info}}

\begin{document}

\maketitle
\begin{abstract}
This paper presents StoryDB — a broad multi-language dataset of narratives. StoryDB is a corpus of texts that includes stories in 42 different languages. Every language includes 500+ stories. Some of the languages include more than 20 000 stories. Every story is indexed across languages and labeled with tags such as a genre or a topic. The corpus shows rich topical and language variation and can serve as a resource for the study of the role of narrative in natural language processing across various languages including low resource ones. We also demonstrate how the dataset could be used to benchmark three modern multilanguage models, namely, mDistillBERT, mBERT, and XLM-RoBERTa.
\end{abstract}

\section{Introduction}

Stories are central to human culture and communication. However, it seems that stories are easier said than generated. Despite incredible recent progress in natural language processing generation of longer texts is still a challenge \cite{van2019narrative, rashkin2020plotmachines}. \citet{ostermann2019mcscript2} present a machine comprehension corpus for the end-to-end evaluation of script knowledge with 50\%  of the questions in the corpus that require script knowledge for the correct answer. The authors demonstrate that though the task is not challenging to humans, existing machine comprehension models fail to perform well on the data, even if they make use of a commonsense knowledge base. 

Partially, this challenge could be attributed to the lack of adequate memory models. Longer texts demand better memory mechanisms and possible ways to construct such mechanisms are discussed in the literature for the last 25 years. Long short-term memory networks \cite{hochreiter1997long}, Neural Turing Machines \cite{graves2014neural}, memory networks \cite{weston2014memory} and many other architectures try to tackle this problem. Attempts to introduce some form of memory in transformers, such as \cite{guo2019star} or \cite{burtsev2020memory}, could be regarded as the next steps in this long line of work. 

 There are some interesting recent attempts to generate long texts using some form that makes such longer text feasible for a human reader. For example, \citet{agafonova2020paranoid} generate a diary of a neural network. Yet the generation of a narrative is still challenging. For a detailed review of earlier approaches to narrative generation, we address the reader to \cite{kybartas2016survey}. Even modern models for narrative generation rely heavily on some form of expert knowledge or some type of hierarchical structure of the narrative. For example, \citet{fan2019strategies} first generate the predicate-argument structure of the text, then generate a surface realization of the predicate-argument structure, finally replace the entity placeholders with context-sensitive names and references. \citet{fan2019strategies,ammanabrolu2020story} propose a hierarchical generation framework that first plans a storyline, and then generate a story based on it.  present a technique for preprocessing textual story data into event sequences. \citet{xu2018skeleton} develop a model that first generates the most critical phrases, called skeleton, and then expands the skeleton to a complete and fluent sentence. Similarly, \citet{martin2018dungeons} provide a mid-level of abstraction between words and a sentence to minimize event sparsity and present a technique for automated story generation whereby the problem is decomposed into the generation of successive events and the generation of natural language sentences from events. Finally, \citet{brahman2020cue} develop an approach, where the user provides the model with such mid-level sentence abstractions in the form of cue phrases during the generation process.

However, we should take into consideration that modern Natural Language Processing (NLP) is fundamentally an experimental discipline, so the lack of dedicated data could be another bottleneck for the development of narrative generation. This paper tries to amend this problem.

Unfortunately, the majority of available narrative datasets deal with some constrained form of a short plot that is usually called {\em scenario}. These scenarios are centered around common activities, i.e. going grocery shopping or taking a shower. These narrative datasets available in the literature are also extremely small and could not be used with the most advanced modern NLP models. \citet{regneri2010learning} collect  493 event sequence descriptions for 22 behavior scenarios. \citet{modi2016inscript} present InScript dataset that consists of 1,000 stories centered around 10 different scenarios.  \citet{wanzare2019detecting} provide 200 scenarios and attempt to identify all references to them in a collection of narrative texts. \citet{mostafazadeh2016corpus} present a corpus of 50k five-sentence commonsense stories. Finally, there is an MPST dataset that contains 14K movie plot synopses, \cite{kar2018mpst}, and WikiPlots\footnote{https://github.com/markriedl/WikiPlots} that contains 112 936 story plots extracted from English language Wikipedia. Recently \citet{malyshevaDYP} provided a dataset of TV series along with an instrument for narrative arc analysis. These datasets are useful yet as well as a vast majority of the narrative datasets they are only available in English.

This paper provides a large multi-language dataset of stories in natural language.  The stories have a cross-language index and every story and character are cross-linked if they occur in different languages. Additionally, the texts have tags such as a genre or a topic. This is the first story dataset of such magnitude that we know of.  We hope that a large dataset of long storylines could be used for various aspects of narrative research as well as to facilitate experiments with end-to-end narrative generation.

\section{Data}

StoryDB is motivated by several interesting experiments that used WikiPlots — one of the larger English datasets of narratives available for all-purpose narrative research that we have mentioned earlier. Seeing various applications that Wikiplots dataset found in the NLP community, we believe, that StoryDB would be even more useful due to multiple languages, advanced filtering that guarantees higher quality of obtained data, and genre tagging. To improve reproducibility and make StoryDB usable as Wikipedia is further updated we publish the data as well as the code for the filtering pipeline\footnote{https://drive.google.com/drive/folders/ 1RCWk7pyvIpubtsf-f2pIsfqTkvtV80Yv}. The stories that form StoryDB are extracted from any Wikipedia article that contains a sub-header that contains the word "plot" (e.g., "Plot", "Plot Summary", etc.) in a corresponding language. 

\subsection{Dataset structure}

The dataset consists of several index files and includes a directory \verb"plots". Every file in the directory has a similar structure. Two first letters of the filename stand for the ISO 639-1\footnote{https://en.wikipedia.org/wiki/ISO\_639-1} code of the language for the texts presented in the file. For example, \verb"hy_plots.tsv" contains 4 861 plots in Armenian language. The file \verb"simple_plots.tsv" contains stories in Simple English. Every entry in the plots file has a similar structure and includes the following fields: 

\begin{itemize}
    \item ID — the unique number of a plot that is the same across every language in the dataset;
    \item Lang — the language of this particular entry;
    \item Link — a link to the Wikipedia page containing the plot;
    \item Title	— the title of the story;
    \item Text — the text of the story;
    \item Categories — the categories that Wikipedia assigns to this story.
\end{itemize}

One can navigate across plot files using StoryDB's Index file \verb"plot_matrix.tsv". The rows of the file stand for languages. If a given plot is available in a given language then the title of this plot stands in the corresponding cell of the \verb"plot_matrix.tsv". For example, if "Wee free men" is available in Simple English it could be found by its title in the corresponding  \verb"simple_plots.tsv". StoryDB also includes \verb"plot_rake.tsv" that contains keywords extracted with RAKE algorithm \cite{rose2010automatic} for every story.

Finally, the files \verb"ID_lang_tag.tsv" and \verb"ID_tag_average.tsv" include information about tags that correspond to the given story. We discuss tagging procedure in detail later.

\subsection{Preprocessing}

Our motivation is to provide a dataset of storylines for various languages including the low-resource ones. Roughly speaking, we want to be sure that every story that ends up in StoryDB is a legitimate storyline description in the corresponding natural language. Thus we are more interested in the precision of the dataset rather than in the recall. To guarantee a higher quality of the obtained stories we implemented several heuristical filters that we briefly describe here. 

English Wikipedia is an order of magnitude bigger than any other Wikipedia both in terms of users and in terms of admins\footnote{https://meta.wikimedia.org/wiki/List\_of\_Wikipedias}. This makes the English list of storylines to be the most extensive one. We regard it as the least noisy one and use it as a reference source for the filtering procedure. We exclude every page that includes a plot yet has no plot section in English Wikipedia for the same entry. 

If Wikipedia in language X has a page with title A and this page is also available in language Y under title B, we list such pair of stories as \verb"[language_X," \verb"title_A," \verb"language_Y," \verb"title_B"]". Every entry in this list is an edge in a graph of stories. Every vertex in this graph has a corresponding name \verb"language, title". Unlike connected stories from different languages that usually contain similar storylines, the stories listed under the same name in the same language might differ significantly. Say, two stories in language X \verb"[language_X," \verb"title_A]" and \verb"[language_X," \verb"title_B]" are both linked to one story in another language Y \verb"[language_Y," \verb"title_C]". To avoid such ambiguities we exclude fully connected components that contain more than one entry in the same language. Obtained list of stories ends up in the resulting matrix of stories to navigate the dataset. We experimented with various filtering procedures and found this combination to produce a sufficiently rich dataset with a minimum amount of duplicates.

StoryDB is also equipped with a catalog of characters. If a given character that has an individual Wikipedia page is mentioned in a story, its description in the original language is saved into the corresponding tsv-file alongside the ID of a story and the language of the description.

\subsection{Tagging}

We annotate the resulting stories using meta-information on categories from Wiki API\footnote{https://www.mediawiki.org/w/api.php?action=help\&
modules=query\%2Bcategories}. For every plot, we list all translated categories marked in every language in which this plot is available. Then we search these category lists for substrings that include tags from the manually created list of tags\footnote{The list of tags is published with the dataset.}. This allows us to proved language-specific tags for every language, that are listed in  \verb"ID_lang_tag.tsv". For example, Czech version of Black Night has tags \verb"action;" \verb"crime;" \verb"drama;" \verb"superhero;" \verb"comics;" and \verb"thriller", while the same story in Persian has no tag  \verb"comics", but has additional tags  \verb"neo-noir;"  \verb"psychological;" \verb"epic;" and \verb"screenplays;".

\begin{figure}[t!]\centering
     \includegraphics[scale=0.2]{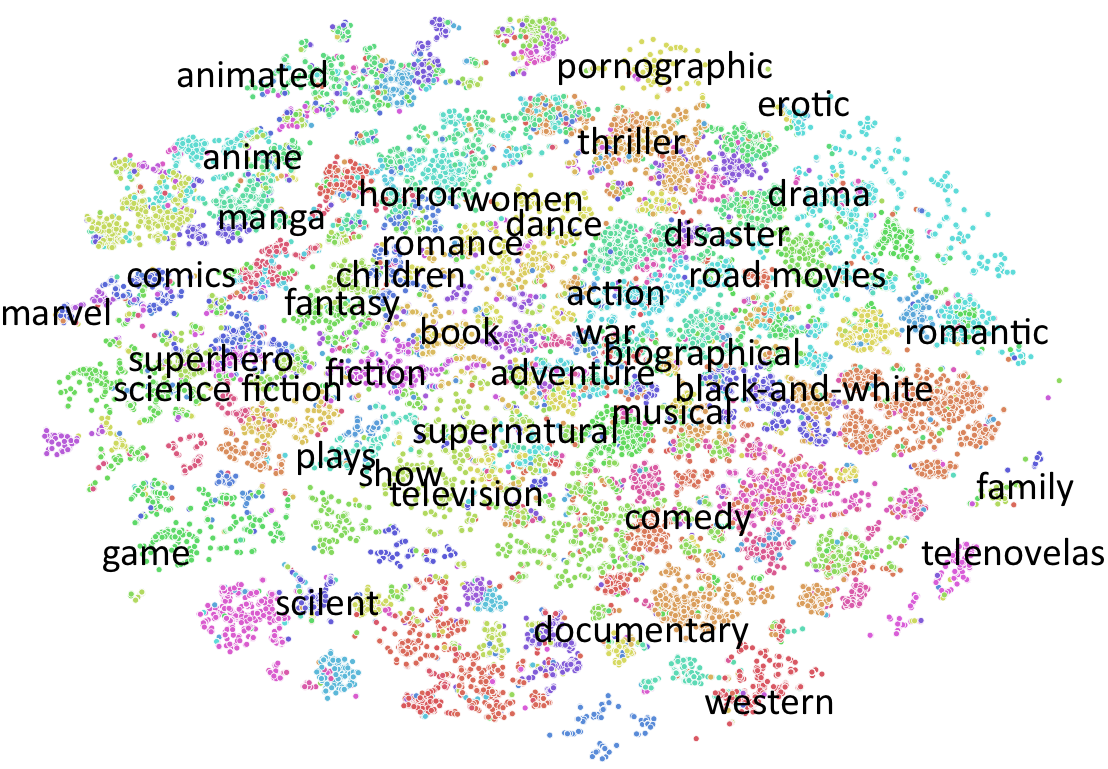}
  \caption{t-SNE visualisation for plots in StoryDB clustered according to their tags. Figure shows centroids of the tags with higher variance across the dataset.}
  \label{fig:tags}
\end{figure}

File \verb"ID_tag_average.tsv" includes the scores of the tags available for every story. The scores are calculated as follows: we count the number of times that a given tag is associated with a given story. Then we divide this number over the total number of languages in which the story is represented. The obtained space of tags could be useful for narrative exploration. Every story becomes a vector with every coordinate on the interval $[0,1]$. Figure \ref{fig:tags} shows a t-SNE visualisation of this space \cite{van2008visualizing} alongside the centroids of the more distinctive tags. 

\subsection{StoryDB}

\begin{figure*}[h!]\centering
     \includegraphics[scale=0.2]{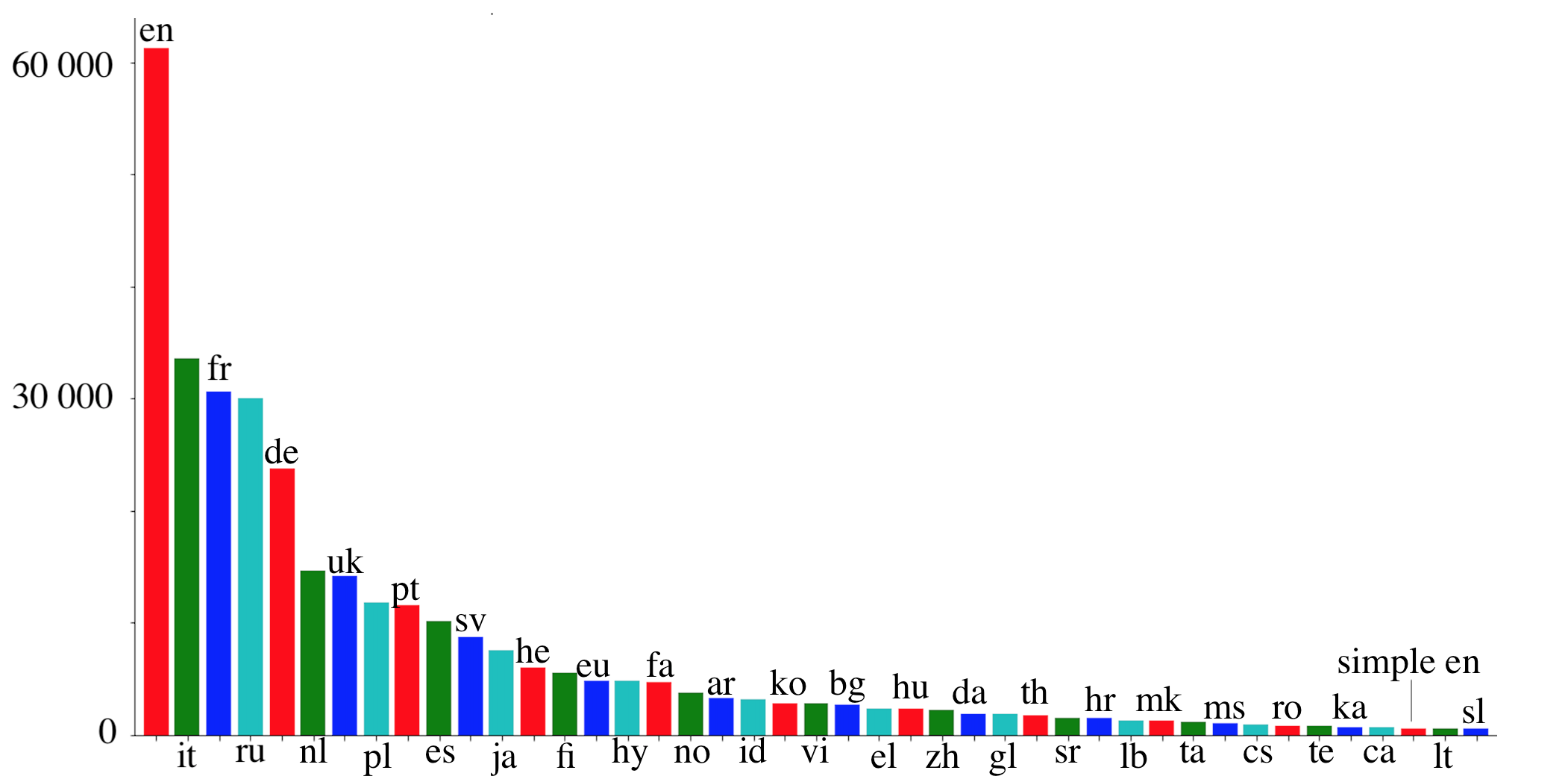}
  \caption{Number of stories in every language that has more that five hundred entries in StoryDB.}
  \label{fig:size}
\end{figure*}

Figure \ref{fig:size} shows the relative size of the datasets in every language presented in StoryDB. English heavily dominates followed by Italian, French, Russian, and German. 

There are more than 20 languages that have three thousand or more stories available, including such languages as Finnish, Hungarian, or Persian. Table \ref{tab:DB} summarises some of the resulting parameters of the obtained dataset.

\begin{table}[h!]
\centering
\begin{tabular}{lr}
\multicolumn{2}{c}{Story DB}                                                                \\ \hline
Number of languages                                                                & 42     \\
\begin{tabular}[c]{@{}l@{}}Median \# of stories in a language\end{tabular}   & 2 772  \\
\begin{tabular}[c]{@{}l@{}}Maximal \# of stories in a language\end{tabular} & 63 756 \\
\begin{tabular}[c]{@{}l@{}}Minimum \# of stories in a language\end{tabular} & 568   
\end{tabular}
  \caption{Some resulting parameters of the StoryDB.}
  \label{tab:DB}
\end{table}

\section{Evaluation}

We have used three modern transformer-based architectures for the evaluation:
\begin{itemize}
    \item mBERT\footnote{https://huggingface.co/bert-base-multilingual-cased} \cite{devlin2018bert} — a multi-language version of BERT;
    \item mDistilBERT\footnote{https://huggingface.co/distilbert-base-multilingual-cased} \cite{Sanh2019DistilBERTAD} — a distilled version of multi-language BERT;
    \item XLM-Roberta\footnote{https://huggingface.co/xlm-roberta-base} \cite{conneau2020unsupervised} — a model that is two times larger than BERT in terms of the number of parameters.
\end{itemize}
 
 These models are the most widely used multi-language models to date. The results of the experiments are publicly available at  Weights and Biases\footnote{https://wandb.ai/altsoph/storydb
\_eval.task1\\ https://wandb.ai/altsoph/storydb\_eval.task3 \\ https://wandb.ai/altsoph/storydb\_eval.task3}, see \cite{wandb}. The evaluation was performed on ten largest languages in StoryDB, namely: English — 'en', French — 'fr', Italian — 'it', Russian — 'ru', German — 'de', Dutch — 'nl', Ukrainian — 'uk', Portuguese — 'pt', Polish — 'pl', and Spanish — 'es'.
 
 We evaluated three tasks:
 \begin{itemize}
    \item Task A. Multilabel classification for tags on a multilanguage corpus of plots;
    \item Task B. Multilabel classification for tags in cross-lingual learning;
    \item Task C. Multilabel classification for tags in cross-lingual learning with a corpus of overlapping plots that occur in every language.
\end{itemize}
 
 Let us now describe every task in detail.
 
 \subsection{Task A}
 
 We have sampled the ten most frequents tags out of StoryDB (tag 'film' was the most frequent yet was excluded as a somewhat redundant one). These tags were: 'drama', 'comedy', 'television', 'fiction', 'series', 'action',
 'thriller', 'black-and-white', 'science fiction', 'horror'. These ten tags form a vector, where every dimension corresponds to one particular tag. '1' encodes the presence of the tag and '0' stands for the absence of it.

For every language out of the top ten in StoryDB, we have sampled 2000 plots such that every plot has at least one tag out of the list of the ten most popular tags. In Task A the plots were sampled randomly for every language, so there is some overlap between languages. On average, 2\% of the plots in one language reoccur in another one. It is important to note that the set of tags for a given plot might differ across languages and one plot could have several tags simultaneously. Thus, multilabel classification is a natural evaluation task under these circumstances. 

Since the dataset is not balanced with respect to tags we used the binary cross-entropy loss\footnote{https://pytorch.org/docs/stable/generated/ torch.nn.BCELoss.html} over the vector of tags. Table \ref{tab:task_a} and Table \ref{tab:task_a_detail} sum up the results of three models on a multilanguage dataset of plots. Further details across languages and tags are available online\footnote{https://wandb.ai/altsoph/storydb\_eval.task1}.

\begin{table}[]
\begin{tabular}{lll}
             & \begin{tabular}[c]{@{}l@{}}Hamming\\ Score\end{tabular} & \begin{tabular}[c]{@{}l@{}}Multilabel\\ Accuracy\end{tabular} \\
             \hline
mDistillBERT & 0.47                                                    & 0.31                                                          \\
mBERT        & 0.50                                                    & 0.33                                                          \\
XLM-RoBERTa  & 0.50                                                    & 0.33                                                         
\end{tabular}
\caption{Task A. Hamming score and multilabel accuracy for the vector of predicted tags on a validation set. Training data consists of sixteen thousand plots in ten languages, with one tenth of the dataset in every language.}
\label{tab:task_a}
\end{table}

\begin{table}[]
\begin{tabular}{llll}
                                                           & mD-BERT & mBERT & XLM-R \\
\hline
Comedy                                                     & 0.69         & 0.67 & 0.69        \\
Action                                                     & 0.67         & 0.70 & 0.67        \\
Fiction                                                    & 0.78         & 0.80 & 0.81        \\
Thriller                                                   & 0.67         & 0.63 & 0.64        \\
Horror                                                     & 0.70         & 0.76 & 0.75        \\
Drama                                                      & 0.73         & 0.74 & 0.74        \\
Series                                                     & 0.77         & 0.78 & 0.78        \\
Television                                                 & 0.74         & 0.76 & 0.76        \\
\begin{tabular}[c]{@{}l@{}}Science\\ Fiction\end{tabular}  & 0.78         & 0.80 & 0.81        \\
\begin{tabular}[c]{@{}l@{}}Black and \\ White\end{tabular} & 0.68         & 0.65 & 0.62       
\end{tabular}
\caption{Task A. AUC-ROC for binary tag classifiers on a validation set. Training data consists of sixteen thousand plots in ten languages, with one tenth of the dataset in every language.}
\label{tab:task_a_detail}
\end{table}

 \subsection{Task B}
 
 Now let us do a similar setup yet train every model on one language in StoryDB and test its accuracy on another language. The parameters of the training datasets and labels are the same as in Task A above but every model is trained on one dataset and is then tested on other languages. Table \ref{tab:task_b} show the performance of mBERT, yet mDistillBERT and XLM-RoBERTa demonstrate similar behavior. The detailed results could be found online\footnote{https://wandb.ai/altsoph/storydb\_eval.task2}.
 
 \begin{table*}[]
 \centering
\begin{tabular}{l|llllllllll}
   & en   & de   & nl   & fr   & it   & es   & pt   & ru   & uk   & pl   \\
   \hline
en &    0.36 & 0.16 & 0.14 & 0.16 & 0.10 & 0.17 & 0.15 & 0.13 & 0.12 & 0.15 \\
de & 0.15 & 0.40 & 0.16 & 0.18 & 0.12 & 0.16 & 0.20  & 0.19 & 0.18 & 0.21 \\
nl & 0.20 & 0.33 & 0.41 & 0.20 & 0.22 & 0.32 & 0.29 & 0.31 & 0.25 & 0.30 \\
fr & 0.16 & 0.20 & 0.16 & 0.51 & 0.13 & 0.18 & 0.18 & 0.16 & 0.14 & 0.19 \\
it & 0.19 & 0.30 & 0.24 & 0.21 & 0.21 & 0.26 & 0.28 & 0.27 & 0.24 & 0.30 \\
es & 0.23 & 0.24 & 0.20 & 0.22 & 0.18 & 0.45 & 0.27 & 0.22 & 0.22 & 0.23 \\
pt & 0.15 & 0.21 & 0.17 & 0.22 & 0.10 & 0.19 & 0.44 & 0.19 & 0.14 & 0.23 \\
ru & 0.12 & 0.21 & 0.16 & 0.13 & 0.12 & 0.20 & 0.22 & 0.45 & 0.16 & 0.20 \\
uk & 0.10 & 0.20 & 0.16 & 0.14 & 0.09 & 0.19 & 0.19 & 0.23 & 0.25 & 0.20 \\
pl & 0.19 & 0.27 & 0.19 & 0.21 & 0.11 & 0.20 & 0.24 & 0.24 & 0.20 & 0.48
\end{tabular}
\caption{Task B. Multilabel accuracy for the vector of predicted tags by mBERT. Training data consists of one thousand six hundred plots in one language. Every row shows validation accuracy of a model pretrained on the corresponding language and validated on the plots in a language from the corresponding column.}
\label{tab:task_b}
\end{table*}
 
 Table \ref{tab:task_b} demonstrates that if we train the model on one language and validate it on the other the quality of the multilabel tag classification drops. This drop varies across languages and tends to be smaller for the languages that belong to the same language family.
 
 \subsection{Task C}
 
 The last validation is similar to Task B, yet now we sample plots that overlap in every language. This limits us to 1500 plots in six languages that we split into train and test. Now every plot occurs in every language. Table \ref{tab:task_c} shows the model manages to recover certain tags in one language after the pre-training on the other. Table \ref{tab:task_c} shows the performance of XLM-RoBERTa, yet mDistillBERT and mBERT demonstrate similar behavior. The performance of the models tends to be better on overlapping plots if we compare it to Task B. The detailed results could be found online\footnote{https://wandb.ai/altsoph/storydb\_eval.task3}.
 
  \begin{table}[]
\begin{tabular}{l|llllll}
   & en   & de   & nl   & fr   & it   & es    \\
   \hline
en & 0.29 & 0.16 & 0.12 & 0.12 & 0.10 & 0.08 \\
de & 0.27 & 0.31 & 0.18 & 0.19 & 0.15 & 0.11 \\
nl & 0.34 & 0.30 & 0.32 & 0.17 & 0.13 & 0.17 \\
fr & 0.22 & 0.20 & 0.16 & 0.27 & 0.15 & 0.10 \\
it & 0.34 & 0.29 & 0.23 & 0.25 & 0.25 & 0.16 \\
ru & 0.25 & 0.21 & 0.18 & 0.18 & 0.18 & 0.13
\end{tabular}
\caption{Task C. Multilabel accuracy for the vector of predicted tags by XLM-RoBERTa across the dataset of plots withour cross-language overlaps. Training data consists of one thousand two hundred plots in one language. Every row shows validation accuracy of a model pretrained on the corresponding language and validated on the plots in a language from the corresponding column.}
\label{tab:task_c}
\end{table}

The multilabel accuracy for tag prediction declines further yet it can neither be attributed to specific lexical properties of a particular language nor any form of overlap of plots across languages. 

This series of evaluation tasks demonstrates two crucial properties of StoryDB:

\begin{itemize}
\item StoryDB could be used to work with narrative structures on the most abstract cross-lingual level;
\item StoryDB allows controlling for various cross-lingual similarities of plots during ablation experiments with models of narrative.
\end{itemize}

\section{Discussion}

We believe that a broad multilanguage dataset of narratives can facilitate several areas of narrative research. 

\begin{itemize}
    \item Cross-cultural research of narrative structure. StoryDB provides possibilities to compare the structure of narrative in various languages. Since StoryDB includes every story in its original language and is equipped with a universal system of tags it is a natural source for such cross-cultural research.
    \item Classification of narratives. StoryDB includes an extensive amount of narratives for various languages alongside their genre tags. This allows to develop new methods for narrative classification as well as extensively test the ones that already exist, see for example \cite{reiter2014nlp}.
    \item Quantitative research of the narrative structure. \cite{y2007employing} represents a story as a cluster of emotional links and tensions between characters that progress over storytime. StoryDB includes the description of the plots alongside the key characters. Such information could be insightful for a deeper quantitative understanding of narrative as a by-product of character interaction. 
    \item Summarization of narrative. Parallel corpora in different languages contain similar descriptions of the narrative that could vary in terms of details and length. That makes StoryDB a useful tool for potential narrative summarization research such as \cite{barros2019natsum}. 
    \item End-to-end narrative generation. StoryDB is the first dataset of narratives that we know of that contains narrative descriptions in various natural languages.
\end{itemize}

\section{Conclusion}

This paper presents StoryDB — a broad multi-language dataset of narratives. We describe the construction of the dataset, provide the code for the whole pipeline, list the parameters of the resulting dataset, and briefly discuss several areas of natural language processing research, where StoryDB could be useful for the community.

We hope that StoryDB could be broadened as more plot descriptions are added to various languages. These considerations make StoryDB a flexible resource that would be relevant for the NLP community as the subfield of quantitative narrative research moves on.

\bibliographystyle{acl_natbib}
\bibliography{storydb}

\end{document}